\documentclass[pdflatex,sn-mathphys-num]{sn-jnl}
\usepackage{graphicx}%
\usepackage{multirow}%
\usepackage{amsmath,amssymb,amsfonts}%
\usepackage{amsthm}%
\usepackage{mathrsfs}%
\usepackage[title]{appendix}%
\usepackage{xcolor}%
\usepackage{textcomp}%
\usepackage{manyfoot}%
\usepackage{booktabs}%
\usepackage{algorithm}%
\usepackage{algorithmicx}%
\usepackage{algpseudocode}%
\usepackage{listings}%
\usepackage{color,soul}%
\usepackage{subcaption}%
\usepackage{xcolor}%
\usepackage{hyperref}

\theoremstyle{thmstyleone}%

\theoremstyle{thmstyletwo}%

\theoremstyle{thmstylethree}%

\begin{document}
\title[
Article Title]{Active Learning for Efficient Annotation of Surgical Videos with Weak Supervision}

\author*[1,2]{\fnm{Manasa} \sur{Dendukuri}}\email{maniden@seas.upenn.edu}

\author[1,2]{\fnm{Matja{\v z}} \sur{Jogan}} 

\author[1,2,3]{\fnm{Daniel A.} \sur{Hashimoto}\textsuperscript{$\dagger$} \equalcont{Co-senior authors jointly directed this work.}}

\author*[1,2 ]{\fnm{Guiqiu} \sur{Liao}\textsuperscript{$\dagger$}\equalcont{Co-senior authors jointly directed this work.}}
\email{guiqiu.liao@pennmedicine.upenn.edu}
\affil*[1]{\orgdiv{GRASP Laboratory}, \orgname{University of Pennsylvania}, \orgaddress{\street{3330 Walnut Street}, \city{Philadelphia}, \state{PA}, \postcode{19104}, \country{USA}}}

\affil[2]{\orgdiv{PCASO Laboratory, Dept. of Surgery}, \orgname{University of Pennsylvania}, \orgaddress{\street{3400 Spruce Street}, \city{Philadelphia}, \state{PA}, \postcode{19104}, \country{USA}}}

\affil[3]{\orgdiv{Dept. of Computer and Information Science}, \orgname{University of Pennsylvania}, \orgaddress{\street{3330 Walnut Street}, \city{Philadelphia}, \state{PA}, \postcode{19104}, \country{USA}}}

\abstract{\textbf{Purpose:} Precise spatial-temporal annotation of laparoscopic videos is time-consuming and requires expert knowledge. We propose a human-in-the-loop knowledge acquisition framework that combines active learning with dual-loss optimization to significantly reduce the annotation effort needed for automatic localization and segmentation of objects in the surgical field.

\textbf{Methods}: Our method employs a foundation model to generate temporally consistent class activation maps (CAMs) from video using two complementary training objectives: a weak supervision loss on video-level tool presence labels for weakly annotated data, and an image-level mask loss on human-corrected annotations obtained through active learning. Rather than requiring dense pixel-level annotation upfront, our pipeline iteratively proposes pseudo-masks that guide the expert annotator to refine the knowledge previously captured by the model.

\textbf{Results}: We demonstrate that our framework reduces the effort of surgical video annotation by 50\% by the end of training in comparison to fully manual annotation.

\textbf{Conclusion}: Through eliminating the need for large, fully annotated datasets from the start, this framework enables scalability to the
development of surgical tool segmentation models. This iterative human-in-the-loop refinement supports efficient knowledge acquisition with minimal expert input, providing a practical and deployable strategy for expanding tool segmentation to larger, more diverse datasets and real-world clinical settings.
Code shared at: \href{https://github.com/PCASOlab/AL-WSL}{https://github.com/PCASOlab/AL-WSL}.
}


\maketitle
\section{Introduction}\label{sec1}

Human labeling is still at the core of machine learning, especially for domain-specific language and computer vision models in high-stakes settings such as healthcare. Surgical video represents one of the richest untapped data sources in healthcare. With the tremendous number of laparoscopic and robotic procedures performed annually worldwide, each generating hours of high-value footage, there is significant potential for AI-driven surgical analysis. However, pixel-level annotation of surgical scenes requires extensive expert time and domain knowledge, creating a critical bottleneck for developing computer-assisted surgical systems.

Traditional supervised and semi-supervised approaches for surgical video segmentation \cite{twinanda2017endonet, zhao2020learning, Lou2023-rq} require dense, frame-wise pixel-level annotations of object classes such as surgical instruments, organs, and tissue, making them impractical for large-scale adoption. Weakly supervised learning (WSL) emerged as a promising alternative, leveraging much cheaper video- or image-level class presence labels to generate class activation maps (CAMs) \cite{zhou2016learning}. In surgical domains, WSL approaches \cite{li2021learning,liao2025disentangling} and fully self-supervised approaches~\cite{Liao2026-kv} have demonstrated feasibility, but reaching the performance needed for semantic segmentation in real-world deployment will require iterative refinement with human feedback, as well as scalable pipelines for ingestion of new data and continuous model adaptation.

Active Learning (AL), or Human-in-the-Loop annotation, integrates human feedback by strategically selecting data for annotation, maximizing model performance while minimizing labeling cost \cite{settles2009active}. Query strategies include uncertainty sampling \cite{lewis1994sequential}, diversity-based selection \cite{sener2018active}, and hybrid approaches \cite{ash2020deep}. In medical imaging, active learning has reduced annotation requirements for segmentation tasks \cite{nath2020diminishing, yang2017suggestive}. Related work in object detection offers complementary insights. Previously proposed active learning strategies for object detection vary in supervision and acquisition criteria. Active Teacher \citep{pengactiveteacher} extends teacher‑student learning by iteratively selecting unlabeled images based on difficulty, information, and diversity, achieving full supervised performance with far fewer labels. In weakly supervised object detection (WSOD), Box-in-Box (BiB) \citep{huyeccv22} improves object localization by selecting uncertain predictions for targeted bounding-box annotation, reducing errors caused by the model focusing on only parts of an object. ALWOD \citep{wang2023alwod} combines weak and active learning: it bootstraps a warm‑start generator, then uses student–teacher disagreement to select informative samples, and reformulates annotation as a rapid verification task to cut labeling hours. While these methods excel on natural image benchmarks (e.g., MS‑COCO), they focus on detection (bounding boxes) and require bounding‑box or fully‑labeled inputs. However, integration of human input with outputs from a weakly supervised learning framework that iteratively learns using weak annotations and pixel-level feedback remains underexplored. 

We propose an expert-in-the-loop framework that synergistically combines weak supervision with active learning to minimize annotation effort while achieving robust surgical tool segmentation and allowing for extrapolation to any visible surgical targets. Our method employs a DINOv3 \cite{simeoni2025dinov3} foundation model with temporal consistency networks \cite{liao2025disentangling} to generate CAMs from video sequences, where the model is trained through dual-loss optimization: a weak supervision loss using only video-level labels for data without mask annotations, and a mask supervision loss incorporating sparse mask annotations obtained as human corrections of model predictions via active learning. Unlike prior work, our approach requires no initial dense annotations and operates through iterative refinement cycles where the annotator corrects only a set amount of randomly selected videos. 

Our contributions are threefold: (1) a novel WSL framework integrating foundation representations with temporal modeling for surgical tool segmentation; (2) an AL pipeline that solicits sparse human feedback to correct and improve predictions; and (3) comprehensive experiments on Cholec80 demonstrating that our approach achieves performance competitive with fully supervised methods while requiring significantly less annotation time (around 50\% reduction). This work represents a significant step toward overcoming the annotation bottleneck in surgical video analysis.

\section{Method}\label{sec2}

\subsection{Overall pipeline}

\begin{figure}[h]
    \centering
    \includegraphics[width=\textwidth]{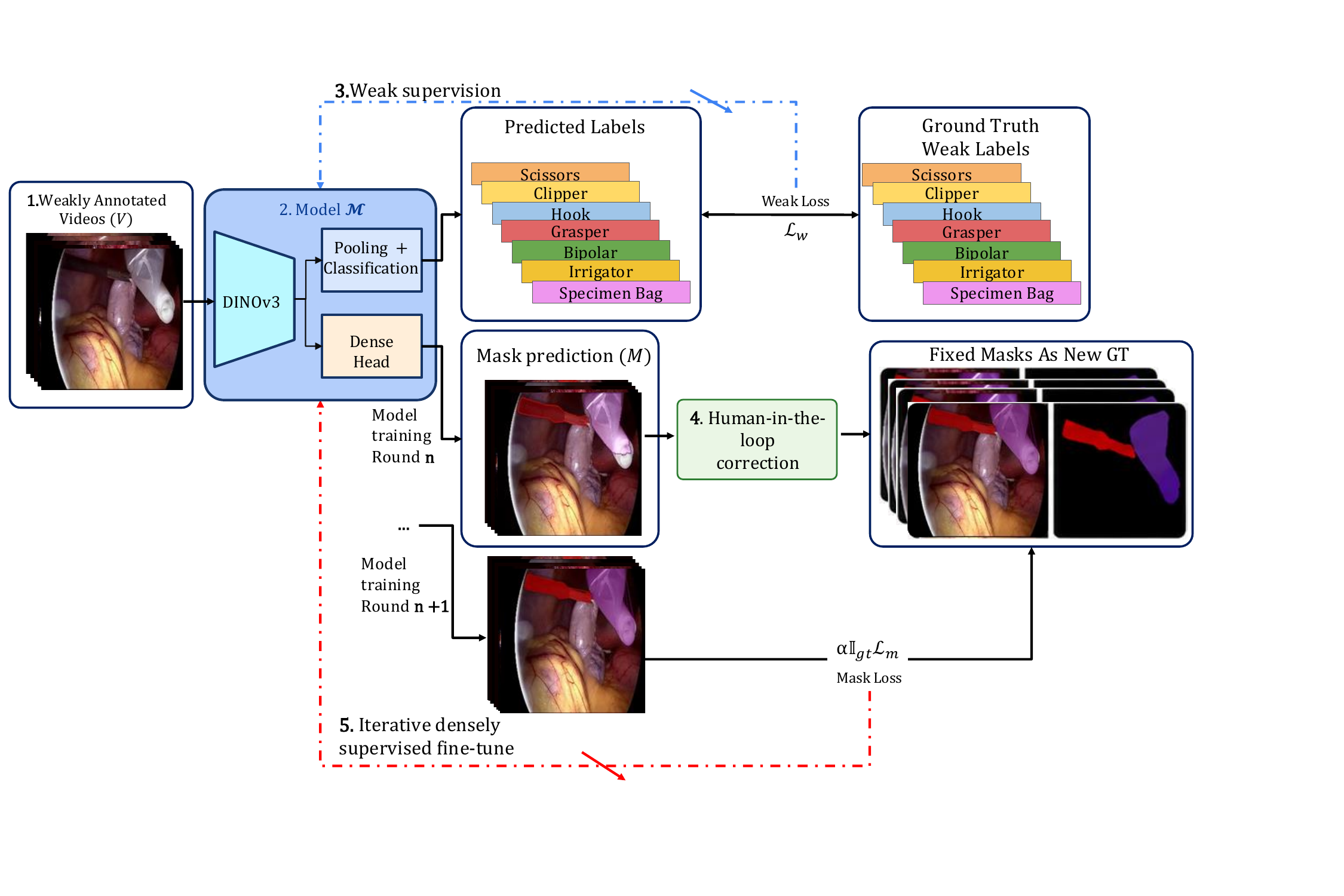}
    \caption{Architecture of the overall pipeline. Videos are sent into Model $\mathcal{M}$, which predicts masks and weak labels. Predicted masks are sent to an online video annotation platform, where a human annotator will fix the masks of each frame of the video, and a mask loss (fixed vs. predicted) is calculated and added to the weak loss, where fixed masks then become ground truth for the remainder of training.}
    \label{fig:model_diagram}
\end{figure}

Our proposed pipeline for learning pixel-level dense knowledge with minimal human expert input is shown in Figure \ref{fig:model_diagram}. Our method leverages a weak supervision model at the video-level \cite{liao2025disentangling} that operates on frame representations extracted from the DINOv3 ViT-B/16 (Vision Transformer Base with 16×16 patches) foundation model, pretrained in a self-supervised manner on the LVD-1689M dataset, a collection of around 1.68 billion images drawn from a pool of 17 billion public web images \cite{simeoni2025dinov3}. 

We chose DINOv3 over other backbones as evidence showing that it can outperform several popular foundation models (e.g., CLIP and SAM) in downstream dense prediction tasks that lie outside the pretraining domain of those foundation models \cite{heinrich2025radiov2,simeoni2025dinov3,cheng2025changedino}. These frame-level features are aggregated across time and used to jointly predict a video-level classification label for weak supervision and a dense, pixel-level localization map of the tools throughout the clip, without requiring any pixel-level annotations.

Model $\mathcal{M}$ consists of a DINOv3 feature extractor followed by a trainable temporal consistency network. Given a raw video clip $V \in \mathbb{R}^{W \times H \times T}$, each frame is passed independently through DINOv3 to obtain per-frame spatial feature maps. These features are then aggregated across time by the temporal consistency network. This network is implemented as a 4-layer 3D Convolutional Neural Network (3D-CNN) which supports temporal reasoning~\cite{tran2015learning,funke2019using}. Each of the four layers applies 3D convolution kernels of size $3 \times 3 \times 3$ with padding and stride of 1, preserving both the temporal depth and the spatial resolution across layers, followed by batch normalization and a ReLU non-linearity. By convolving jointly across the temporal, height, and width dimensions, the 3D-CNN learns to model motion patterns and temporal co-occurrences among visual features across adjacent frames, augmenting the frame-wise, spatially rich DINOv3 representations.

Throughout training, the DINOv3 backbone remains frozen to preserve its self-supervised representations, while optimization is confined to the temporal consistency and prediction heads. $\mathcal{M}$ has two prediction heads: (1) a video-level classification head that predicts clip-level tool presence labels for weak supervision, and (2) a dense prediction head that produces a class activation map (CAM) for the whole video. The CAMs are computed from the weights of the final fully connected layer of the video-level classification head, projected onto the spatial feature maps produced by the last convolutional block of the temporal consistency network. This layer is selected since its feature maps preserve the highest spatial resolution among all layers while encoding task-relevant semantic information. 

Projecting classification weights onto these maps highlights the most discriminative regions for each tool category, enabling weakly supervised localization without pixel-level annotations. These CAMs are then refined into binary masks using dense Conditional Random Fields (CRFs) \cite{krahenbuhl2011efficient}, yielding masks of size $W \times H \times T \times C$, where $C$ is the number of tool classes. When the model needs to be updated, human annotators receive selected videos and the model's predicted annotations and provide corrections through an online video annotation platform. Once human correction is completed, it gets incorporated into the next rounds of training as ground truth masks. The workflow is interactive and allows annotators to label a flexible number of videos per round, with the proportion of mask-supervised training data growing incrementally as active learning progresses.

\subsection{Active Learning}
\subsubsection{Training protocol}
Training begins with video-level weak labels and classification loss (i.e., weak supervision loss) \cite{liao2025disentangling} and no ground truth segmentation masks. Once a certain number of epochs (\textit{N} = 11 in our experiments) is reached, active learning rounds start running every \textit{M} = 10 epochs. At each AL round, videos are randomly selected for annotation using a sampling probability that is biased toward balancing the minority classes. Predictions from the model are exported for expert correction, and corrected masks are reintegrated into training as ground truth masks. The model is then able to utilize a dual loss: weak supervision loss (applied to all videos) and mask loss (applied only to annotated videos, with a weight $\alpha$,
\begin{equation}
\mathcal{L}_{\text{total}}
=
\mathcal{L}_{\text{w}}
+
\alpha \mathbb{I}_{\text{gt}} \, \mathcal{L}_{\text{m}} ,
\end{equation}
where $\mathbb{I}_{\text{gt}}= [0,1]$ indicates whether the ground truth mask exists at the pixel level. With each round of active learning, the number of ground truth masks available for training increases. 

\subsubsection{Annotation protocol with expert-in-the-loop}
\begin{figure}[h]
    \centering
    
    \begin{subfigure}{0.45\textwidth}
        \centering
        \includegraphics[width=\textwidth]{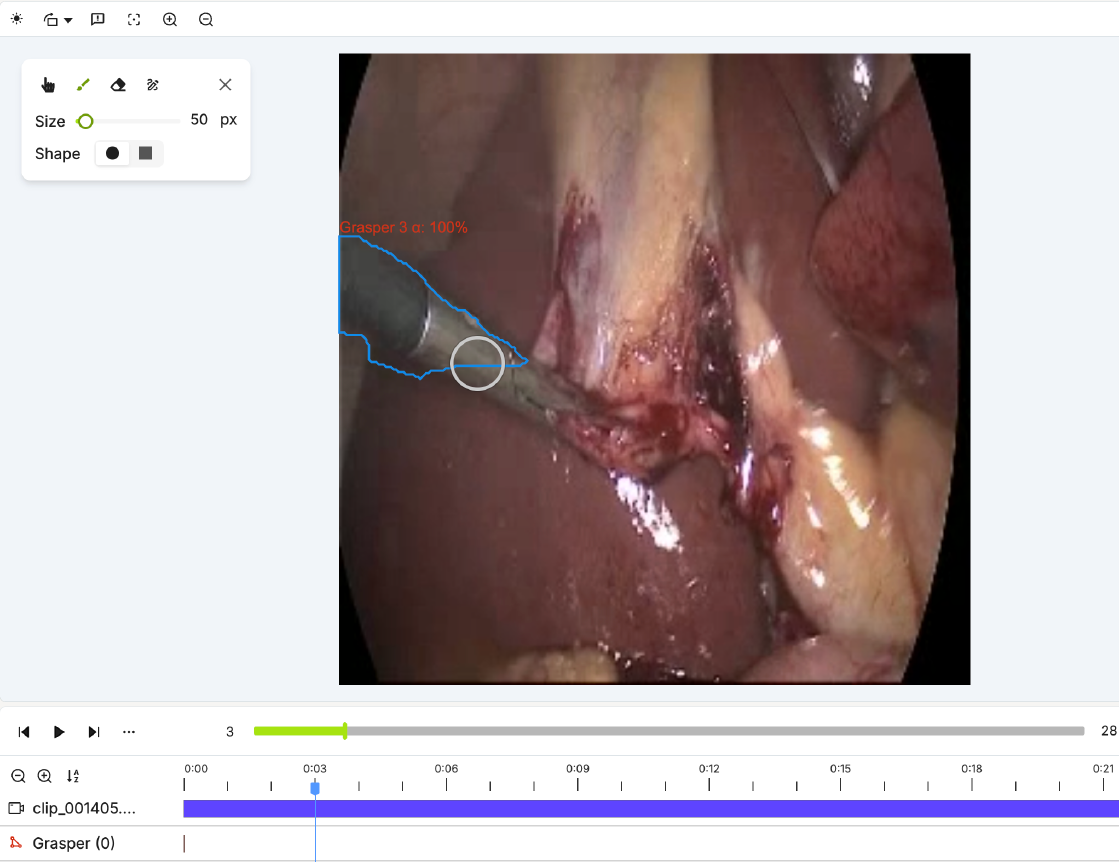}
        \caption{Brush Tool.}
        \label{fig:enc1}
    \end{subfigure}
    \hfill
    \begin{subfigure}{0.45\textwidth}
        \centering
        \includegraphics[width=\textwidth]{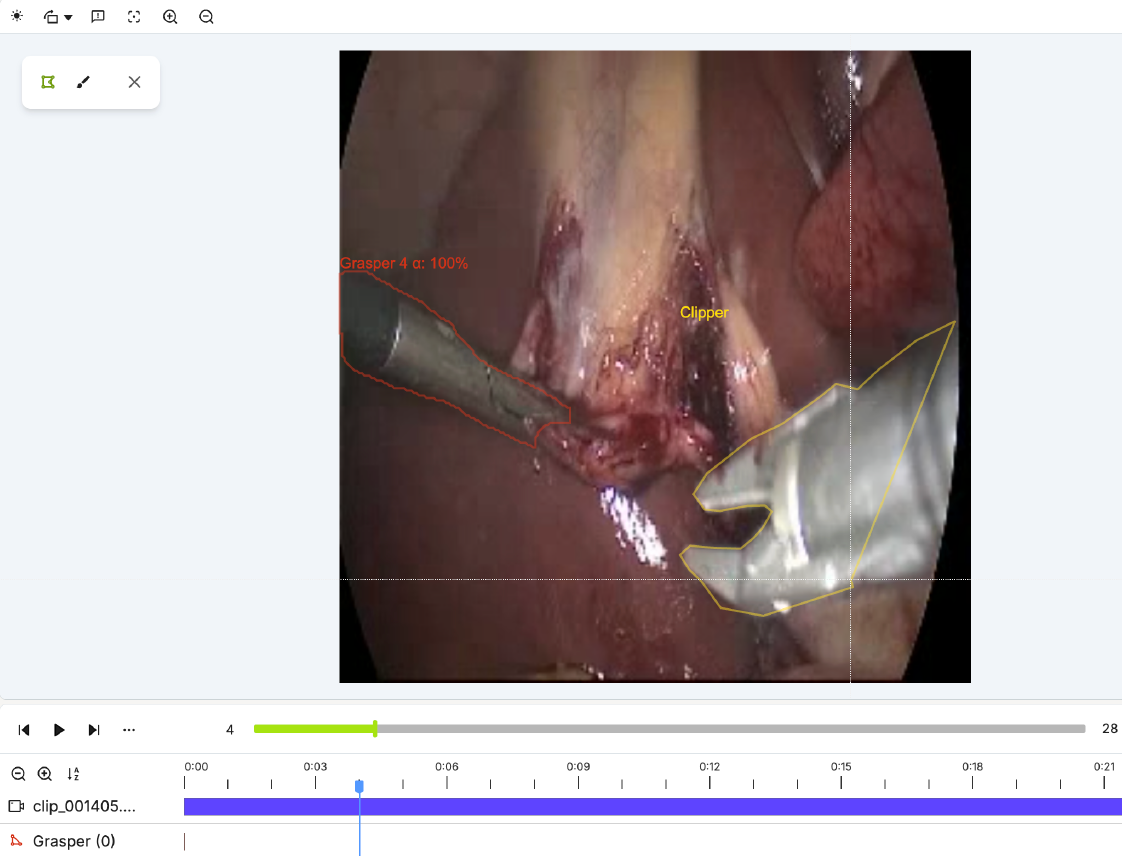}
        \caption{Polygon tool.}
    \end{subfigure}
    
    \caption{Correction of annotations using brush and polygon tool in Encord.}
    \label{fig:encordpic}
\end{figure}

At each active learning round, model-predicted segmentations are exported to the online video annotation platform Encord \cite{encord2026} where human annotators can correct predictions using a polygon or a brush tool to segment surgical tools accurately, as shown in Figure \ref{fig:encordpic}. SAM2 and tracking \cite{ravi2024sam}, while embedded in Encord's workflow as an annotation aid, are not used in our experimental workflow in order to preserve the annotations predicted by the model. We use these tools as a comparative annotation protocol instead.

\section{Experiment and results}\label{sec3}
\subsection{Data}
All experiments used the Cholec80 \cite{twinanda2017endonet} dataset with weak labels for seven surgical tools: Grasper, Bipolar, Hook, Scissors, Clipper, Irrigator, and Specimen Bag. Each video clip consisted of 29 frames, and training was performed on 6,100 clips. The scores and accuracy were found from running all checkpoint rounds through a test set of 81 Cholec80 video clips, which the model was not trained on. 
\subsection{Experiment setup and evaluation objectives}
Three active learning trials were conducted with different annotation budgets per round: \textbf{AL-2} (2 videos per round, 14 total), \textbf{AL-5} (5 videos per round, 35 total), and \textbf{AL-10} (10 videos per round, 70 total). To evaluate the contribution of active learning and weak supervision, a control group (\textbf{Control-WSL}) was tested using the same dual-loss model architecture but with all ground truth masks provided at epoch 0, effectively implementing semi-supervised segmentation with weak supervision. Three variants were evaluated using 14, 35, and 70 annotated videos, matching the final annotation totals of AL-2, AL-5, and AL-10. For clarity we denote these comparisons with the AL equivalent (eg., CWS-10 for 70 total annotated videos). We also test performance with two different mask loss weights ($\alpha \in \{0.01, 0.1\} $). 

A total of 84 video clips from the Cholec80 dataset were annotated from scratch. To simulate real-world variability in annotation times, 35 videos were annotated using the annotation platform's embedded SAM with tracking, another 35 using SAM without tracking, and 14 videos were annotated completely manually without the use of tracking or SAM. This design enables three key comparisons: active learning versus upfront annotation (AL trials vs. Control-WSL, both in time and performance), the effect of annotation budget (14 vs. 35 vs. 70 videos), and the effect of mask loss weight $\alpha$ (0.01 vs. 0.1).

Each experimental model was evaluated on the test dataset over five independent runs. For each run, the Correct Localization (CorLoc) score was recorded, and we report the mean and 95\% bootstrap confidence interval. Additionally, we computed paired differences between AL and CWS models using a BCa bootstrap ($N$=10000 bootstraps on the test set) to quantify the statistical significance of performance differences. Annotation times were recorded using the time logs provided by the annotation platform, where the start task was considered the start time, and the submit task was considered the end time. Timings were also measured using a stopwatch to ensure accuracy of recording annotation time. For each annotation method, a t-test was performed to calculate the statistical difference between the average time spent annotating a single video.
\subsubsection{Annotation Efficiency}
Annotations were performed by two trained annotators with experience in surgical video analysis. One annotator had annotated surgical tools for other datasets, the same tools in different anatomy, while the other was trained to annotate tools by the first annotator in a similar style. Both annotators were also trained to be familiar with the annotation platform before the experimental benchmark. In addition, to ensure reduced biased during annotations, standardized written instructions for tool labeling, mask refinement, and correction procedures were defined and followed. All videos were processed in similar conditions and the same two annotators performed all methods (SAM, SAM2, Manual) to keep annotator skill constant. This dual annotator design balances annotation reliability with practical constraints and reflects realistic human-in-the-loop workflows while maintaining fair comparisons across methods. This design was intentionally adopted to ensure consistency across annotation rounds and to isolate the effect of the proposed active learning pipeline.

Three annotation workflows were evaluated as baselines for comparison with the proposed active learning methods. All annotation experiments were conducted using the same two trained annotators following a fixed annotation protocol and all videos were processed under the same ordering and time-tracking conditions. For SAM, annotations were performed on a frame-by-frame basis, where segmentation masks for each surgical tool were initialized and refined using the interactive SAM interface at every frame. For SAM2, we leveraged its temporal propagation capabilities by initializing segmentation masks on the first frame of each video and propagating labels across subsequent frames. When propagation errors occurred due to occlusions, rapid tool motion, or mask drift, annotators corrected only the affected frames by re-initializing or locally refining masks within the same SAM2 workflow, rather than restarting the full sequence. Manual annotations were performed using polygon and brush tools in a frame-by-frame manner, applying the same labeling guidelines and quality checks as the other workflows. All three methods were evaluated on the same set of videos.
\begin{table}[h]
\centering
\caption{Comparison of annotation methods}
\label{tab:annotation_methods}
\begin{tabular}{rccc}
\hline
\textbf{Method} & \textbf{Number} & \textbf{Total Time} & \textbf{Average Time} \\
 & \textbf{Videos} & \textbf{(min)} & \textbf{per Video (min)} \\
\hline
SAM2 (with tracking)\cite{ravi2024sam} & 35 & 56 & 1.60 \\
SAM without tracking \cite{kirillov2023segment} & 35 & 174 & 4.97 \\
Manual Segmentation & 13 & 82 & 6.34 \\
\hline
AL-2 & 14 & 49 & 3.50 \\
AL-5 & 35 & 168 & 4.80 \\
AL-10 & 70 & 249 & 3.56 \\
\hline
\end{tabular}
\end{table}

\begin{figure}[h]
    \centering
    \includegraphics[width=0.95\textwidth]{ 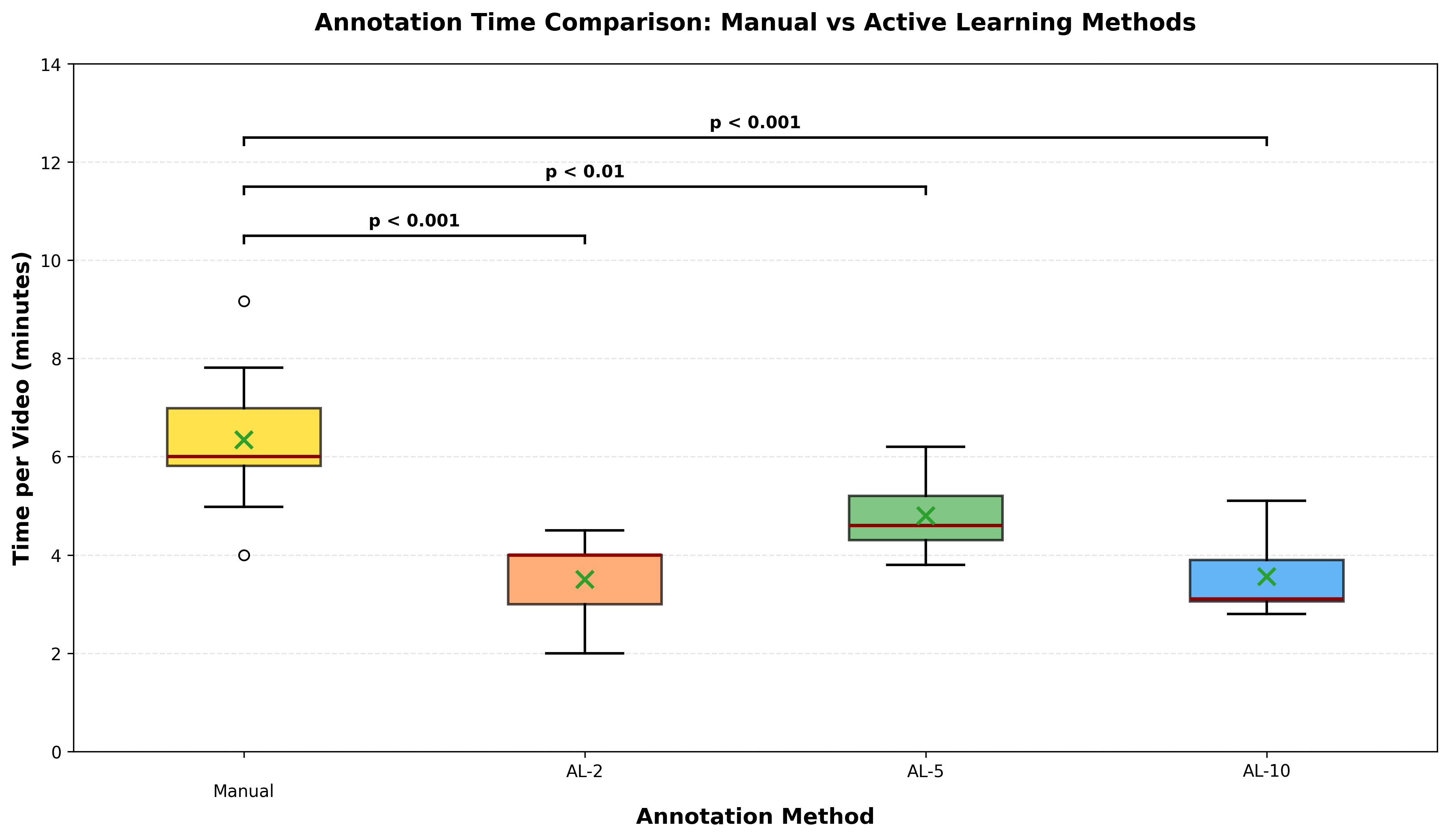}
    \caption{Annotation time comparison across all segmentation methods: complete manual, and active learning with different dataset sizes; all active learning methods take significantly less time compared to manual segmentation approaches }
    \label{fig:boxplot_all_methods}
\end{figure}
The effort put into annotating videos was measured by the amount of time it took to annotate each video. For each active learning (AL) and manual segmentation round, the start time and end time were recorded for each video and were averaged per round. 

Annotation times can be seen in Table \ref{tab:annotation_methods} and Figure \ref{fig:boxplot_all_methods}, which both reveal significant differences in annotation efficiency across methods. Manual segmentation approaches required more time per video than the AL pipeline (Manual vs AL-2: $p<$0.001; Manual vs AL-5: $p<$0.01; Manual vs AL-10: $p<$0.001). The manual segmentation required 6.34 minutes per video, as this method did not use SAM or SAM2 and thus was directly compared to the outcomes of the AL experimental runs.  
The AL approach improved efficiency relative to manual annotations, while achieving performance comparable to the SAM segmentations done manually on the annotation platform. The average time taken to annotate each video for all active learning trials was around 3.953 minutes, less than annotating with just SAM with no tracking. This indicates that iterative active learning not only accelerates the annotation process but does so without sacrificing model performance.

\subsubsection{Performance of Active Learning}
Table \ref{tab:combined_corloc_weights} presents a direct comparison of CorLoc scores between CWS and AL methods across all surgical tool categories. We compare against CWS (the same dual‑loss model trained with all ground truth masks upfront) rather than against prior active learning, as those methods are designed for object detection (bounding boxes) on natural images and require bounding‑box annotations or fully‑labeled initial sets. 
In other words, while prior active learning work focuses on detection with bounding boxes on natural images, our method is tailored to the segmentation of surgical tools from video using only weak video‑level labels.
 
AL consistently matched or exceeded CWS performance while requiring substantially less annotation effort. AL-10 ($\alpha$ = 0.01) achieved an overall average at 0.439, followed by AL-10 ($\alpha$ = 0.1) at 0.428, both significantly higher than the CWS groups ($p <$ 0.001). At the 10-video annotation budget, AL significantly outperformed CWS under both weight settings: AL-10 ($\alpha$ = 0.01) showed a mean difference of +0.061 (95\% CI [0.058, 0.063]), and AL-10 ($\alpha$ = 0.1) showed a mean difference of +0.033 (95\% CI [0.027, 0.039]). At the 2-video budget, AL-2 ($\alpha$ = 0.1) outperformed CWS-2 with a mean difference of +0.029 (95\% CI [0.023, 0.035]), while AL-2 ($\alpha$ = 0.01) underperformed CWS-2 by $-$0.046 (95\% CI [$-$0.051, $-$0.043]) shown in Table \ref{tab:corloc_comparison}. Differences at the 5-video budget were not statistically significant ($p >$ 0.05), with bootstrap CIs crossing zero under both weight settings.
\begin{table*}
\centering
\caption{CorLoc performance for CWS and AL with different mask loss weights.}
\label{tab:combined_corloc_weights}
\setlength{\tabcolsep}{3pt}
\resizebox{\linewidth}{!}{%
\begin{tabular}{r|c|c|ccccccc}
\hline
\textbf{Method} & \textbf{Weight} & \textbf{Average} & \textbf{Grasper} & \textbf{Bipolar} & \textbf{Hook} & \textbf{Scissors} & \textbf{Clipper} & \textbf{Irrigator} & \textbf{Specimen Bag} \\
\hline
\hline
CWS-2 & 0.1 & 0.392 & 0.419 & \textbf{0.350} & 0.449 & 0.022 & \textbf{0.523} & \textbf{0.479} & \textbf{0.498} \\
AL-2  & 0.1 & \textbf{0.420} & \textbf{0.468} & 0.225 & \textbf{0.484} & \textbf{0.368} & 0.460 & 0.454 & 0.482 \\
\hline
CWS-5 & 0.1 & 0.378 & \textbf{0.467} & 0.203 & 0.447 & 0.135 & \textbf{0.440} & \textbf{0.482} & \textbf{0.473} \\
AL-5  & 0.1 & \textbf{0.380} & 0.450 & \textbf{0.217} & \textbf{0.459} & \textbf{0.322} & 0.331 & 0.411 & 0.467 \\
\hline
CWS-10 & 0.1 & 0.395 & 0.456 & \textbf{0.247} & \textbf{0.475} & 0.214 & 0.445 & 0.446 & 0.481 \\
AL-10  & 0.1 & \textbf{0.428} & \textbf{0.466} & 0.240 & 0.471 & \textbf{0.222} & \textbf{0.543} & \textbf{0.535} & \textbf{0.520} \\
\hline
\hline
CWS-2 & 0.01 & \textbf{0.428} & \textbf{0.494} & \textbf{0.296} & \textbf{0.465} & 0.340 & \textbf{0.521} & 0.382 & \textbf{0.497} \\
AL-2  & 0.01 & 0.382 & 0.436 & 0.145 & 0.438 & \textbf{0.350} & 0.420 & \textbf{0.446} & 0.440 \\
\hline
CWS-5 & 0.01 & 0.402 & 0.424 & \textbf{0.157} & 0.433 & \textbf{0.340} & \textbf{0.564} & \textbf{0.404} & 0.487 \\
AL-5  & 0.01 & \textbf{0.404} & \textbf{0.503} & 0.148 & \textbf{0.459} & 0.309 & 0.520 & 0.394 & \textbf{0.498} \\
\hline
CWS-10 & 0.01 & 0.378 & \textbf{0.500} & 0.202 & \textbf{0.443} & 0.000 & 0.514 & \textbf{0.476} & 0.513 \\
AL-10  & 0.01 & \textbf{0.440} & 0.457 & \textbf{0.351} & 0.441 & \textbf{0.346} & \textbf{0.518} & 0.410 & \textbf{0.554} \\
\hline
\end{tabular}}
\end{table*}

Taking the minority classes into consideration, including Bipolar, Scissors, Clipper, Irrigator, and Specimen Bag, AL showed better results in categories Specimen Bag with AL-10 ($\alpha$ = 0.01) at 0.554 ($p <$ 0.01), Scissors with AL-10 ($\alpha$=0.01) at 0.346 ($p <$ 0.001), Clipper with AL-10 ($\alpha$ = 0.1) having a score of 0.543 ($p <$ 0.001), and Irrigator at 0.535 in AL-10 ($\alpha$ = 0.1) ($p <$ 0.001). Overall, AL trials achieved a higher average minority class score (0.390) compared to the control (0.373) ($p <$ 0.017). 
\begin{table}[ht]
\centering
\caption{Bootstrap Confidence Intervals for AL–CWS CorLoc Differences with Multiple Comparison Correction.}
\label{tab:corloc_comparison}
\begin{tabular}{llcc}
\hline
\textbf{Weight} & \textbf{Comparison} & \textbf{$\Delta$ Mean} & \textbf{95\% CI} \\
\hline
\multirow{3}{*}{0.01}
  & AL2 vs CWS-2   & $-0.0458$ & $[-0.0514,\ -0.0428]$ \,$\checkmark$ \\
  & AL5 vs CWS-5   & $+0.0029$ & $[-0.0043,\ +0.0143]$ \,\textrm{ns} \\
  & AL10 vs CWS-10 & $+0.0614$ & $[+0.0577,\ +0.0632]$ \,$\checkmark$ \\
\hline
\multirow{3}{*}{0.1}
  & AL2 vs CWS-2   & $+0.0286$ & $[+0.0234,\ +0.0347]$ \,$\checkmark$ \\
  & AL5 vs CWS-5   & $+0.0061$ & $[-0.0024,\ +0.0121]$ \,\textrm{ns} \\
  & AL10 vs CWS-10 & $+0.0334$ & $[+0.0273,\ +0.0392]$ \,$\checkmark$ \\
\hline
\multicolumn{4}{l}{$\checkmark$ Significant at $\alpha$ = 0.05 (95$\%$ CI excludes zero).\quad \textrm{ns} = not significant.}
\end{tabular}
\end{table}

\subsubsection{Annotation Time Across Active Learning Rounds}
\begin{figure}[h]
    \centering
    \includegraphics[width=0.95\textwidth]{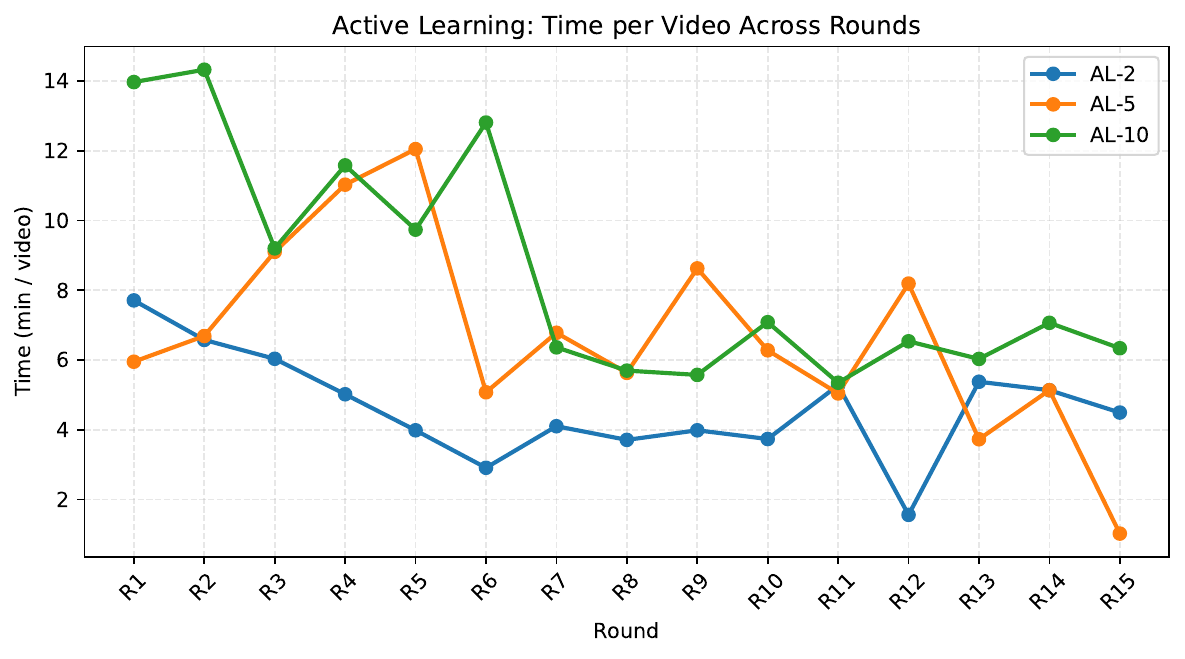}
    \caption{Annotation time per video across active learning rounds for different batch sizes.}
    \label{fig:timeannots}
\end{figure}

The annotation time over multiple AL rounds is shown in Figure \ref{fig:timeannots}. A clear overall downward trend is visible across all three configurations, with annotation times decreasing substantially from early to later rounds, with intermittent fluctuations throughout. For AL-2, annotation time begins at approximately 7.7 minutes per video at R1 and follows a steep decline through to R6 (around 3 minutes), before partially stabilizing in the 3.75-5.5 minute range for the remainder of training, with the exception of a sharp drop to approximately 1.6 minutes at R12. For AL-5, the pattern is somewhat different, where the annotation time initially increases from 6 minutes at R1 to a peak of around 12 minutes at R5, before dropping sharply at R6 (5 minutes) and then oscillating between approximately 4 and 8.5 minutes for the remaining rounds, with a notable drop to 1 minute at R15. AL-10 begins with the highest annotation times of all three configurations, starting at around 14 minutes per video at R1--R2, before declining steeply from R6 onwards and stabilizing around 5.5-7 minutes from R7 through 
to R15.

The initial increase seen in AL-5 between R1 and R5 is likely due to the annotators encountering a run of more complex or instrument-dense videos in the early batches, where the model's weak initialization provides little useful guidance and the annotation burden is therefore high. The subsequent sharp drop at R6 coincides with a point at which the model may have accumulated sufficient training signal to begin producing useful segmentation proposals, allowing annotators to work more efficiently. A similar but less pronounced pattern is visible in AL-10, where early oscillations between R1--R6 give way to a more stable and lower time range from R7 onwards.

\subsubsection{Mask Loss Contribution}
The above results reflect an important dependency on mask loss weight $\alpha$. Across both CWS and AL configurations, $\alpha$ = 0.01 produced stronger overall performance than $\alpha$ = 0.1, particularly at higher annotation budgets. AL-10 improved from 0.428 ($\alpha$ = 0.1) to 0.440 ($\alpha$ = 0.01, $p <$ 0.001), and AL-5 improved from 0.380 ($\alpha$ = 0.1) to 0.404 ($\alpha$ = 0.01, $p <$ 0.01), suggesting that a smaller mask loss contribution allows the model greater flexibility during weakly supervised training, as shown in Table \ref{tab:combined_corloc_weights}. The exception was AL-2, where $\alpha$ = 0.1 (0.420) substantially outperformed $\alpha$ = 0.01 (0.382) ($p <$ 0.001), indicating that with limited annotations, stronger mask supervision provides a useful regularizing effect. This suggests that the optimal mask loss weight is annotation-budget-dependent: smaller budgets benefit from stronger mask guidance, while larger budgets allow the model to learn more freely with reduced mask supervision. Future work will further explore the effect of different selection strategies on mask loss contribution.
\subsubsection{Active Learning Over Multiple Rounds}
\begin{figure}[h]
    \centering
    
    \begin{subfigure}{0.45\textwidth}
        \centering
        \includegraphics[width=\textwidth]{ 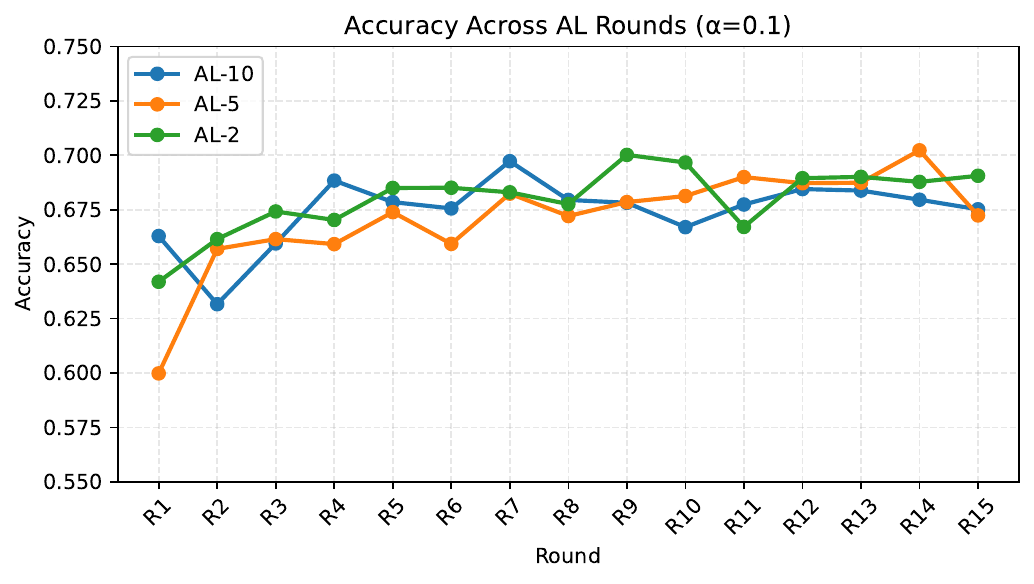}
        \caption{$\alpha=0.1$}
        \label{fig:accur01}
    \end{subfigure}
    \hfill
    \begin{subfigure}{0.45\textwidth}
        \centering
        \includegraphics[width=\textwidth]{ 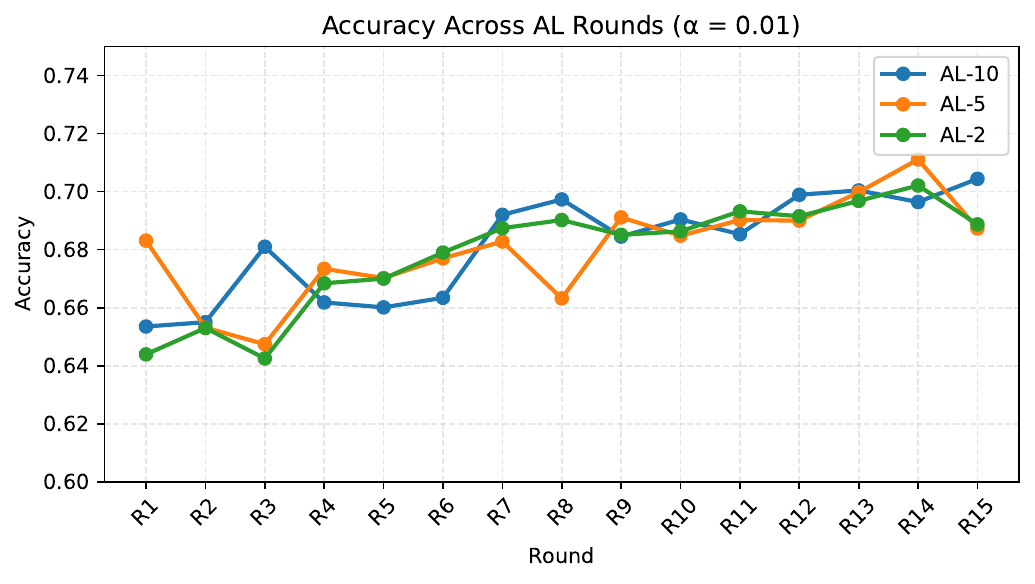}
        \caption{$\alpha=0.01$}
        \label{fig:accur001}
    \end{subfigure}
    
    \caption{Accuracy for different mask loss weights across all AL rounds.}
    \label{fig:accuracies}
\end{figure}

\textbf{Accuracy}
Figure~\ref{fig:accuracies} presents the classification accuracy across 15 active learning rounds under two mask loss weight settings, $\alpha = 0.01$ and $\alpha = 0.1$, for the three experimental settings: AL-2, AL-5, and AL-10. Under both weight conditions, all three configuration overall accuracies increased over the AL rounds, with final accuracies converging in the range of 0.69-0.71. The early rounds show the most variance between configurations: at $\alpha=0.01$, AL-5 begins notably higher than AL-10 and AL-2 at R1, only for AL-10 to overtake it by R3. At $\alpha = 0.1$, AL-10 dips sharply to approximately 0.63 at R2 before recovering, while AL-5 begins at its lowest point of 0.60 before rapidly climbing. These early instabilities reflect the cold-start nature of active learning, where the model has been exposed to relatively few labeled examples and is therefore sensitive to the specific samples selected in each batch. Across both alpha settings, the three configurations largely converge by the mid-rounds (R7-R10), suggesting that batch size has a diminishing influence on accuracy as the labeled pool grows.

\begin{figure}[h]
    \centering

    \begin{subfigure}{0.48\textwidth}
        \centering
        \includegraphics[width=\textwidth]{ 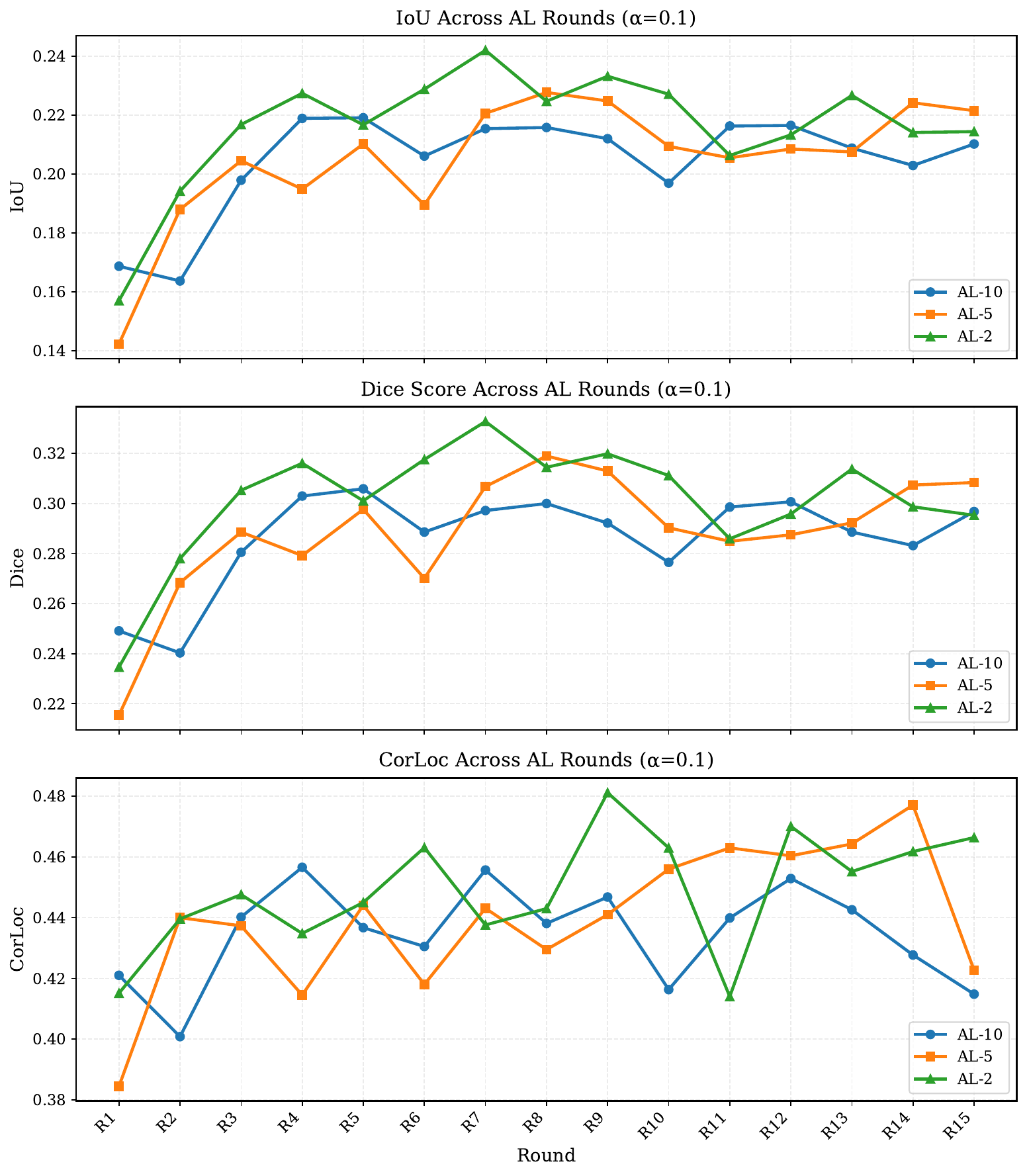}
        \caption{$\alpha$ = 0.1}
        \label{fig:scoresalpha01}
    \end{subfigure}
    \hfill
    \begin{subfigure}{0.48\textwidth}
        \centering
        \includegraphics[width=\textwidth]{ 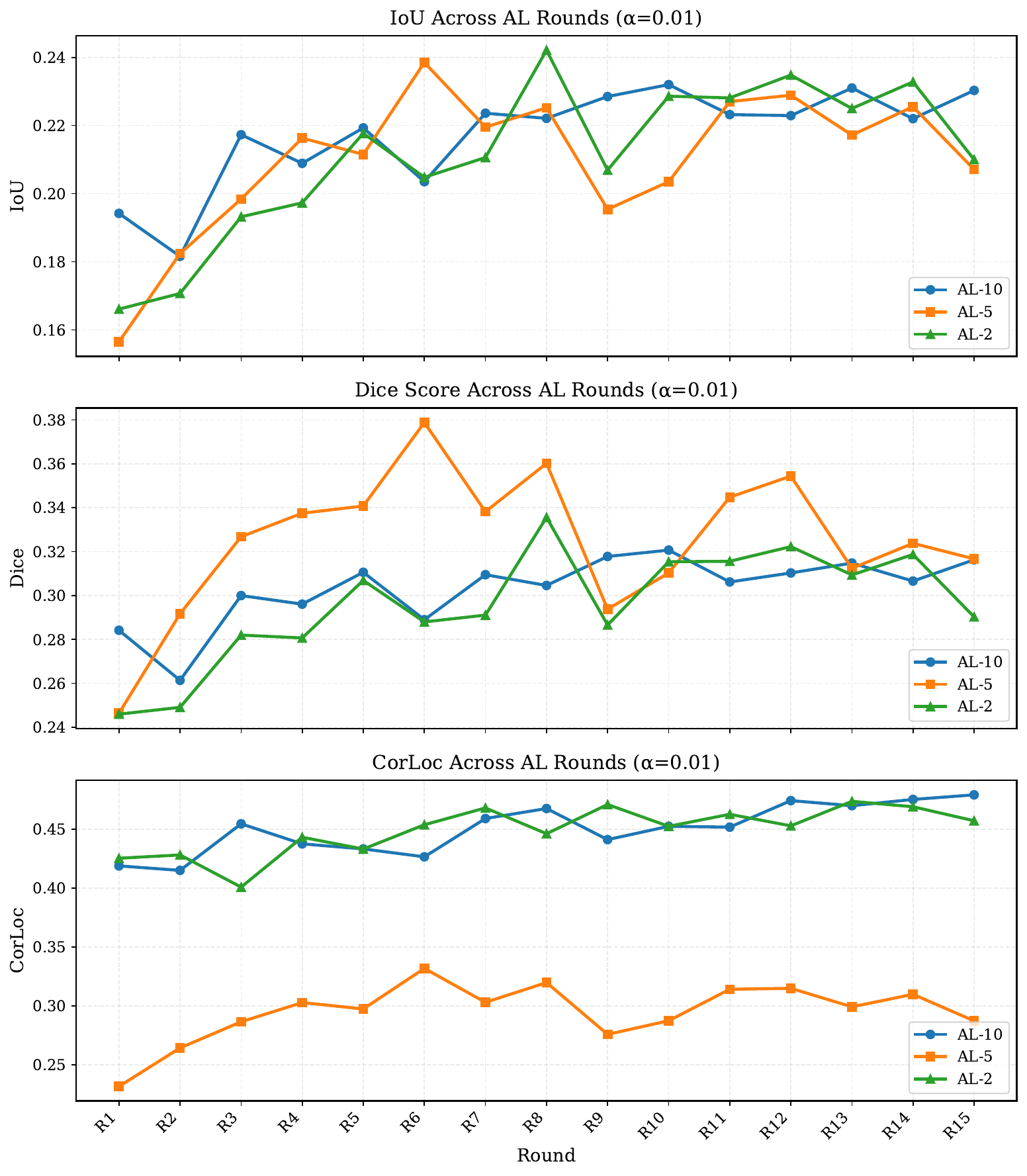}
        \caption{$\alpha$ = 0.01}
        \label{fig:scoresalpha001}
    \end{subfigure}

    \caption{IoU, Dice, Corloc Scores over annotation round checkpoints on testing set.}
    \label{fig:ScoreComparison}
\end{figure}

\textbf{Scores} Figure~\ref{fig:ScoreComparison} reports IoU, Dice score, and CorLoc across 15 active learning rounds under mask loss weights of $\alpha = 0.1$ and $\alpha = 0.01$, respectively. All reported scores are weighted by the population frequency of each surgical tool in the test set. IoU and Dice measure pixel‑level segmentation quality, while CorLoc indicates whether the instrument is correctly localized.

With a smaller mask loss of $\alpha = 0.01$, AL‑10 and AL‑2 start with IoU around $0.19$ and $0.18$ at R1, respectively, and gradually increase to the $0.22$–$0.23$ range from R7 onward. CorLoc reveals a clear divergence: AL‑10 and AL‑2 maintain stable localization scores between $0.42$ and $0.47$, whereas AL‑5 lags at $0.30$–$0.35$ for most rounds. The weak mask signal insufficiently guides AL‑5's smaller per‑round batches toward reliable localization, whereas larger batches in AL‑10 provide enough variety to compensate.

With a stronger mask loss weight ($\alpha = 0.1$, Figure~\ref{fig:ScoreComparison}), the model places greater emphasis on human‑corrected masks when they become available. Smaller, frequent batches under strong mask supervision allow incremental refinement with less disruption. CorLoc under $\alpha = 0.1$ is notably higher for AL‑5 than under $\alpha = 0.01$, with all three budgets reaching $0.44$–$0.48$ by later rounds. The stronger mask signal helps anchor predictions to instrument regions even when batch composition varies, explaining the improved localization consistency.

Across both loss weight settings, IoU, Dice, and CorLoc all exhibit considerable round‑to‑round fluctuation rather than a smooth upward trajectory. Overall, the upward trend visible across all metrics over the full 15 rounds demonstrates that the active learning process is effective, but the granular fluctuations reflect the inherent challenge of learning surgical instrument segmentation from small, randomly composed incremental batches under a weakly supervised mask loss.
\section{Conclusion}\label{sec4}
We presented an active learning pipeline combining weak supervision with dual-loss optimization to reduce annotation effort in surgical video analysis. Our approach achieved a 50\% reduction in annotation time compared to manual segmentation while outperforming upfront annotation controls. Through eliminating the need for large, fully annotated datasets from the start, this framework enables scalability to the development of surgical tool segmentation models. This iterative human-in-the-loop refinement supports efficient knowledge acquisition with minimal expert input, providing a practical and deployable strategy for expanding tool segmentation to larger, more diverse datasets and real-world clinical settings.

\bmhead{Acknowledgments}
This work was supported by the Linda Pechenik Montague Investigator Award and the American Surgical Association Foundation Fellowship.

\bibliography{sn-bibliography,paperpileMJ}
\end{document}